\documentclass[runningheads,orivec]{llncs}
\usepackage{cite}
\usepackage{algorithm}
\usepackage{multirow}
\usepackage{wrapfig}
\usepackage{amsmath,amsfonts}
\usepackage{algorithmic}
\usepackage{graphicx}
\usepackage{textcomp}
\usepackage{xcolor}
\usepackage{footmisc}
\usepackage{booktabs}
\usepackage{paralist}
\usepackage[hyphens]{url}
\usepackage{url}

\usepackage{hyperref} 
\usepackage[caption=false,font=normalsize,labelfont=sf,textfont=sf]{subfig}
\usepackage{orcidlink}

\usepackage{color}

\hypersetup{
  colorlinks=true,
  linkcolor=blue,
  citecolor=blue,
  urlcolor=blue
}

\usepackage{amsmath}
\usepackage{algorithm}
\usepackage{algorithmic}

\usepackage{newfloat}
\usepackage{listings}
\DeclareCaptionStyle{ruled}{labelfont=normalfont,labelsep=colon,strut=off} 
\floatstyle{ruled}
\newfloat{listing}{tb}{lst}{}
\floatname{listing}{Listing}

\begin{document}

\mainmatter 
\title{Synthetic Data in Education: Empirical Insights from Traditional Resampling and Deep Generative Models}
\titlerunning{Synthetic Data in Education} 

\author{
Tapiwa Amion Chinodakufa\inst{1}\orcidlink{0009-0000-3718-3443} \and
Ashfaq Ali Shafin\inst{2}\orcidlink{0000-0002-0135-0091} \and 
Khandaker Mamun Ahmed\inst{1}$^*$\orcidlink{0000-0002-4713-188X}
}
\authorrunning{Chinodakufa et al.} 
\tocauthor{Tapiwa Amion Chinodakufa, Ashfaq Ali Shafin, Khandaker Mamun Ahmed}
\institute{
The Beacom College of Computer \& Cyber Sciences, Dakota State University \and
Knight Foundation School of Computing \& Information Sciences, Florida International University\\
tapi.chinodakufa@trojans.dsu.edu, ashaf016@fiu.edu, khandakermamun.ahmed@dsu.edu
}

\maketitle

\begin{abstract}

Synthetic data generation offers promise for addressing data scarcity and privacy concerns in educational technology, yet practitioners lack empirical guidance for selecting between traditional resampling techniques and modern deep learning approaches. This study presents the first systematic benchmark comparing these paradigms using a 10,000-record student performance dataset. We evaluate three resampling methods (SMOTE, Bootstrap, Random Oversampling) against three deep learning models (Autoencoder, Variational Autoencoder, Copula-GAN) across multiple dimensions: distributional fidelity (Kolmogorov-Smirnov distance, Jensen-Shannon divergence), machine learning utility such as Train-on-Synthetic-Test-on-Real scores (TSTR), and privacy preservation (Distance to Closest Record). Our findings reveal a fundamental trade-off: resampling methods achieve near-perfect utility (TSTR: 0.997) but completely fail privacy protection ( DCR $\approx$ 0.00), while deep learning models provide strong privacy guarantees (DCR $\approx$ 1.00) at significant utility cost. Variational Autoencoders emerge as the optimal compromise, maintaining 83.3\% predictive performance while ensuring complete privacy protection. We also provide actionable recommendations: use traditional resampling for internal development where privacy is controlled, and VAEs for external data sharing where privacy is paramount. This work establishes a foundational benchmark and practical decision framework for synthetic data generation in learning analytics.
\end{abstract}

\section{Introduction}

The field of learning analytics holds significant potential for advancing educational outcomes, yet its development is frequently constrained by limited access to sensitive student data driven by privacy, ethical, and regulatory concerns. Synthetic data generation presents a promising path forward by enabling open research and methodological innovation while reducing reliance on identifiable student information. By producing artificial datasets that preserve the essential statistical characteristics of real data without exposing individual records, synthetic data offers a viable mechanism for supporting data-driven advances in a privacy-respecting manner.

Despite the advancements offered by deep generative models to generate synthetic data, traditional resampling strategies remain widely employed for data augmentation and for addressing class imbalance. However, the literature lacks a direct comparison of their ability to generate holistically realistic tabular data in the educational domain. This study aims to fill that gap by providing a direct benchmark of these two distinct paradigms. Our research question is: 

\noindent
{\bf RQ}. What are the empirical trade-offs between simple, widely accessible resampling methods and complex, computationally intensive deep learning models when generating synthetic educational data?

By answering this question, this paper provides a foundational benchmark for practitioners and researchers, offering clear insights into the strengths and weaknesses of each approach and guiding the selection of the right tool for a given data synthesis task in education.

A recent lawsuit against an EdTech company for allegedly selling K-12 student data without consent~\cite{italiano2024edtech} highlights the challenges of using real-world data to train AI and machine learning models in educational contexts. First, Data scarcity poses a fundamental obstacle as large-scale, high-quality datasets are often unavailable, particularly for rare learning events or specialized educational contexts~\cite{alzubaidi2023survey}. Detailed student performance records typically remain confined to proprietary software systems, inaccessible to the research community. Second, Privacy and ethical constraints further limit data availability. Student records contain personally identifiable information protected by regulations such as FERPA, restricting data sharing even when datasets exist~\cite{parks2017beyond}. Beyond privacy, real-world educational datasets frequently encode historical and systemic biases that can perpetuate inequities in AI-driven learning systems~\cite{baker2021algorithmic,idowu2024debiasing}. The financial and computational costs of acquiring, processing, and securing educational data present additional barriers, particularly for resource-constrained institutions.


 
\section{Related Works}

Recent research explores diverse methods for synthetic data generation, with a central focus on balancing statistical fidelity, privacy protection, and utility across various domains. Differential privacy (DP) has emerged as the gold standard for privacy-preserving synthetic data generation~\cite{perez2024does}. However, recent evaluations reveal critical limitations: DP-synthetic data can exhibit dramatically inflated Type I errors, particularly at stringent privacy levels, necessitating careful validation and substantial original dataset sizes to ensure statistical reliability~\cite{perez2024does}. To address broader machine learning tasks, Bowen et al.~\cite{bowen2019comparative} extend differential privacy through \emph{metric privacy}, enabling private synthetic data generation suitable for clustering and classification applications.

Beyond privacy-preservation, synthetic data addresses domain-specific challenges such as fairness and class imbalance. Jiang et al.~\cite{Jiang024synthetic} propose a genetic algorithm-based method to generate synthetic educational datasets with controllable unfairness properties for benchmarking fairness-aware algorithms. Their approach increases dataset unfairness by 156.3\% on average while preserving predictive model performance, addressing the scarcity of diverse fairness benchmarks in education. For class imbalance, techniques such as SMOTE and ADASYN~\cite{cu2024increment} generate minority-class samples through interpolation or adaptive resampling, enhancing model generalization for underrepresented groups such as at-risk students.

Recent advances extend synthetic data generation beyond tabular formats. The Conditional Probabilistic Auto-Regressive (CPAR) model~\cite{zhang2022sequential} enables generation of sequential data with conditional inputs and custom loss functions, handling mixed data types and irregular temporal intervals. Gaussian Copulas offer another approach by modeling inter-feature dependencies separately from marginal distributions~\cite{combrink2022comparing}. In natural language processing, LLM-generated synthetic data shows promise for benchmarking tasks such as intent detection and text similarity, though performance degrades for complex tasks like named entity recognition~\cite{maheshwari2024efficacy}.

Despite these advances, synthetic data utility remains task-dependent and often falls short of real-world datasets. Lee et al.~\cite{lee2024exploring} propose evaluation metrics including \emph{train2test distance} and \emph{APt2t} to assess representational quality and predictive power, demonstrating that synthetic data enhances model performance primarily under limited real data conditions. Recent work by Khalil et al.~\cite{khalil2025creating} explores synthetic data generation in education with emphasis on privacy preservation and fairness using generative models such as CTGAN and LLMs. While their research prioritizes privacy guarantees and fairness metrics, our work focuses on reconciling statistical fidelity with downstream utility through detailed distributional analysis. Specifically, our model-by-model KDE comparisons, paired with metrics such as JS divergence, TSTR score, and categorical fidelity, offer granular insights into distributional realism and practical guidance for model selection.

Overall, the literature establishes synthetic data as a powerful complement to real data for addressing scarcity, privacy, and bias. However, challenges persist in maintaining distributional fidelity, ensuring differential privacy under tight budgets, and achieving reliable performance across diverse tasks. Our work contributes to this landscape by providing comprehensive evaluation frameworks that bridge theoretical soundness with practical deployment considerations.

\section{Dataset}

\begin{table*}[!t]
\centering
\footnotesize
\caption{Example records from the student performance dataset~\cite{ajeed2023students} used to generate synthetic data, showing mixed categorical-continuous features typical of educational tabular data.}
\label{tab:student_sample10k}
\resizebox{\textwidth}{!}{
\begin{tabular}{@{}lccccccccc@{}}
\textbf{Gender} & 
\textbf{Race/} & 
\textbf{Parental} & 
\textbf{Lunch} & 
\textbf{Test} & 
\textbf{Math} & 
\textbf{Reading} & 
\textbf{Writing} & 
\textbf{Science} & 
\textbf{Total} \\ 
& \textbf{Ethnicity} & \textbf{Education} & & \textbf{Prep}& & & & &\textbf{Score}\\
\hline
Male   & B & High School    & 1 & 0 & 65 & 100 & 67 & 96 & 328 \\
Male   & C & Master's       & 0 & 0 & 10 & 99  & 97 & 58 & 264 \\
Male   & D & Some College   & 1 & 1 & 22 & 51  & 41 & 84 & 198 \\
Female & A & Associate's    & 0 & 0 & 87 & 66  & 76 & 61 & 290 \\
Female & C & Associate's    & 1 & 0 & 62 & 36  & 79 & 63 & 240 \\ 
\hline
\end{tabular}
}
\end{table*}

We used the Student Performance Dataset for Academic Insights \cite{ajeed2023students}, which is publicly available on Kaggle. The dataset comprises records for 10,000 students and is characterized by ten key features: Gender (male, female), race/ethnicity (A, B, C, D, E), parental level of education (some high school, high school, some college, associates, bachelors, masters), free or reduced lunch status (0, 1), completion of a test preparation course (0, 1), and math (0-100), reading (0-100), science (0-100), writing (0-100), total score (0-400). In Table \ref{tab:student_sample10k}, we represent a subset of the original dataset.

\section{Methodology}

\begin{figure*}[!ht]
    \centering
    \begin{minipage}[b]{\textwidth}
        \centering
    \includegraphics[width=\textwidth, keepaspectratio]{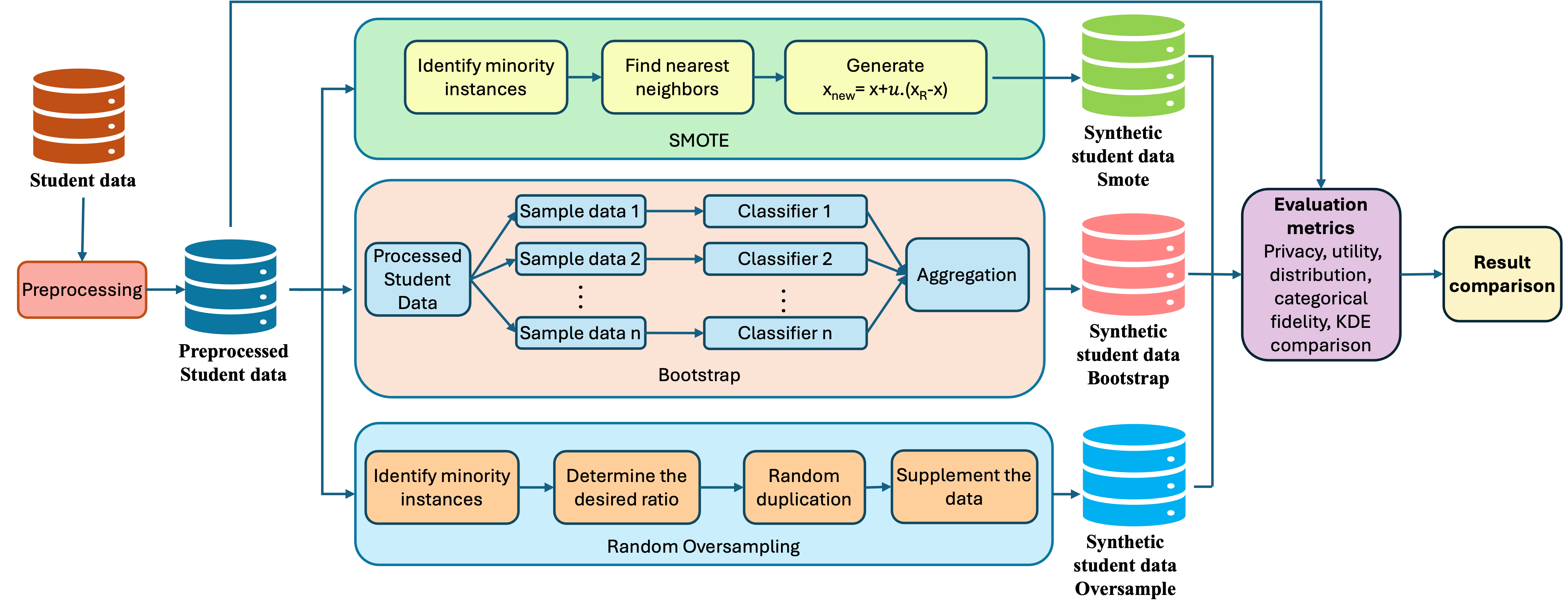}
        {(a) Architecture of traditional methods for synthetic data generation: (1) SMOTE; (2) Bootstrap; and (3) Random oversampling.}
    \end{minipage}
    \hfill
    \vspace{2pt}
    \begin{minipage}[b]{\textwidth}
        \centering
        \includegraphics[width=\textwidth, keepaspectratio]{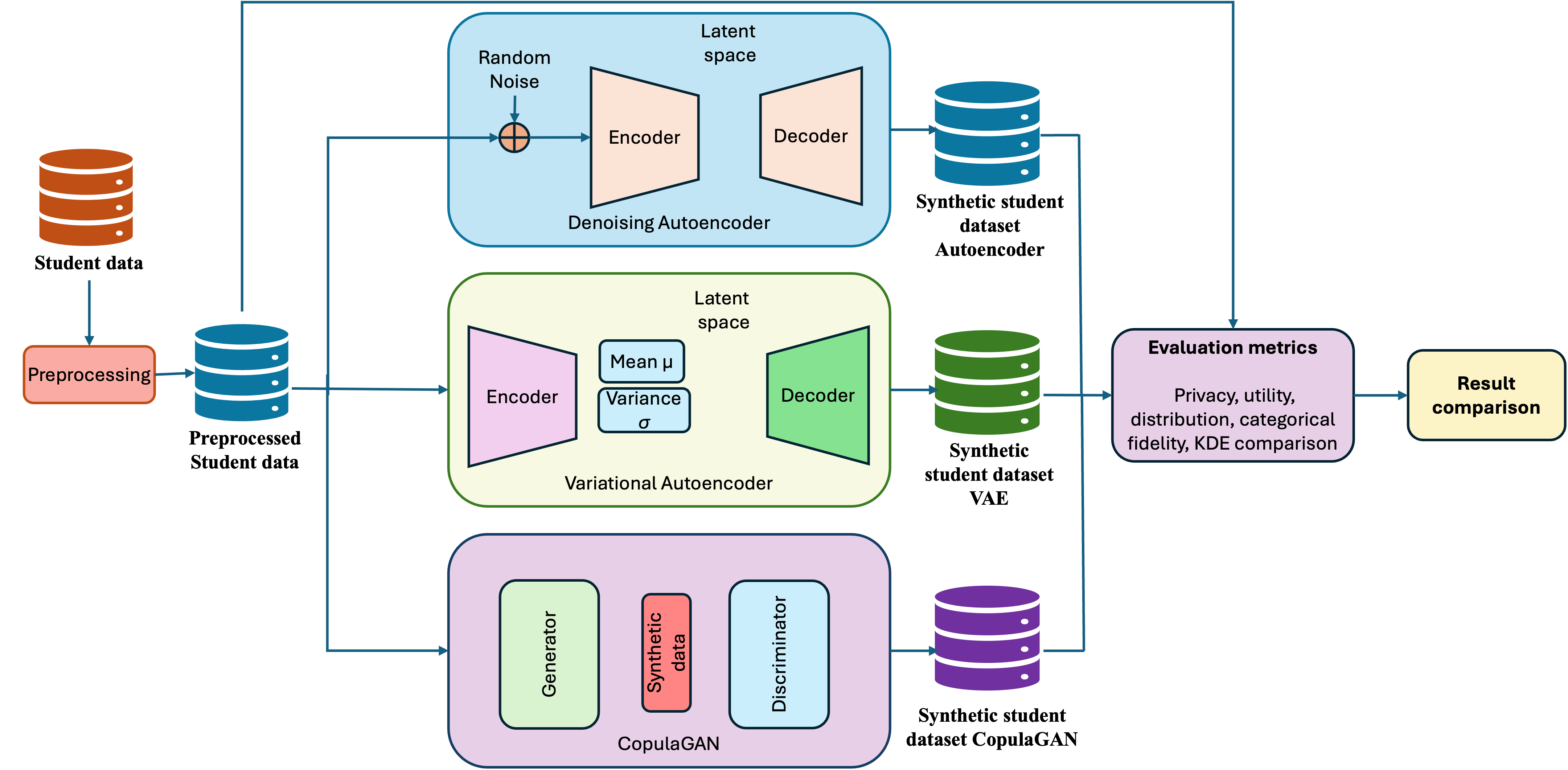}
        {(b) Architecture of deep learning methods for synthetic data generation: (1) Denoising autoencoder; (2) Variational autoencoder; and (3) CopulaGAN.}
    \end{minipage}
    \caption{Architecture of synthetic data generation. In (a), we represent the process of generating synthetic data using traditional methods and in (b), we represent deep learning methods for generating synthetic data.}
    \label{fig:workflow_diagram}
\end{figure*}

Our pipeline proceeds in two main stages. First, in preprocessing step, we preprocess the raw tabular data by applying one-hot encoding to categorical/binary attributes: Gender, Race/Ethnicity, Parental Education Level, Lunch status, and Test Preparation Course participation and converting them into a numerical form suitable for downstream modeling. In this step, we also remove duplicate records and entries with missing values, and standardize continuous variables as needed to promote stable and well-conditioned optimization across models. Second, to mitigate class imbalance and improve the robustness of learned representations, we construct resampled variants of the training set using three complementary strategies: SMOTE to synthesize minority examples via interpolations in feature space, Bootstrap resampling to generate diverse training subsets through sampling with replacement, and Random Oversampling to increase minority-class prevalence without altering feature values. We also train a suite of deep generative models: Autoencoder (AE), Variational Autoencoder (VAE), and Generative Adversarial Network (GAN). This design allows us to compare classical resampling against learned generative augmentation, and to assess which approach yields the most informative synthetic data and downstream predictive gains. For the GAN component, we adopt Copula-GAN \cite{patki2016synthetic, xu2019modeling}, which is designed to better preserve marginal feature distributions and capture inter-feature dependencies. Using this models, we generate 10,000 synthetic student records. We then evaluate all approaches under a common set of metrics and compare their performance. In Figure \ref{fig:workflow_diagram}, we illustrate the architecture of our proposed method that includes the architectural description of each model.

\subsection{Model Implementation}
The preprocessing pipeline was instantiated with race-ethnicity as the target variable for resampling. Accordingly, the feature matrix $X$ comprised all columns except the target, while the label vector $y$ contained only the race-ethnicity attribute. We explicitly specified categorical and numerical feature sets to ensure consistent type-aware processing and compatibility across resampling procedures and generative modeling approaches.

\subsubsection{Resampling Methods.} We consider three standard resampling techniques in our experiment: SMOTE, Bootstrap sampling and Random oversampling which are discussed in this section.

\noindent
{\bf SMOTE}.
The Synthetic Minority Over-sampling Technique (SMOTE)~\cite{chawla2002smote} was implemented with imbalanced-learn, using random\_state=42 and k\_neighbors=5. To support mixed-type inputs, we identified the indices of categorical features and provided them to the categorical features argument, ensuring that non-numeric attributes were treated consistently during interpolation. SMOTE was applied to the original student dataset, which exhibited substantial imbalance in Race/Ethnicity (32\% class C vs. 7\% class A), Lunch status (64\% standard vs. 36\% free/reduced), and Test Preparation Course participation (67\% none vs. 33\% completed). Operationally, SMOTE synthesizes minority-class instances by interpolating between each minority sample and one of its five nearest minority neighbors in feature space. Specifically, for a given minority point, the method selects a neighbor among the $k$ closest minority samples and generates a synthetic instance at a randomly chosen location along the line segment connecting the two points. This yields new samples that are plausible convex combinations of real minority observations, increasing minority density in sparse regions without duplicating existing records.

\noindent
{\bf Bootstrap Sampling}.
Bootstrap resampling was performed using scikit-learn's resample utility. We generated $100$ bootstrap datasets $n_{bootstrap}$ = $100$ with replacement, such that each resampled dataset preserved the original sample size, i.e., ($n_{samples} = |D_{original}|$). Each bootstrap replicate therefore constitutes a stochastic reweighting of the empirical distribution: some student records may be selected multiple times, while others may be omitted in a given replicate. This procedure induces controlled variability across training sets and provides a robust baseline for assessing model stability and performance under data perturbations.

\noindent
{\bf Random Oversampling}.
Random oversampling was performed with imbalanced-learn’s RandomOverSampler. This baseline method addresses class imbalance by uniformly sampling minority-class instances with replacement, effectively duplicating existing records until the target class proportions are reached~\cite{krawczyk2016learning}. While it does not introduce novel samples, it provides a strong and widely used reference point for evaluating more sophisticated resampling and generative approaches.

\subsubsection{Deep Learning Methods.} To generate synthetic data, we used three deep learning based solutions.

\noindent
{\bf VAE}.
Variational Autoencoder (VAE)  instantiates a variational autoencoder tailored to tabular data, learning a stochastic latent representation of the joint feature distribution~\cite{kingma2013auto}. Unlike deterministic autoencoders, the VAE parameterizes each latent dimension as a distribution (typically Gaussian) and samples from these learned posteriors during decoding. VAE method was implemented using the Synthetic Data Vault (SDV) framework~\cite{patki2016synthetic}. We retained SDV’s default hyperparameters for the model implementation where we set the same min/max boundaries of the real data and the enforce rounding is also True. Metadata was also detected from the dataset. The model was trained on the complete dataset via the fit() method, and synthetic data was generated using sample(num\_rows=10,000) to match the original dataset's cardinality using 300 epochs. 

\noindent
{\bf CopulaGAN}.
CopulaGAN adapts the conventional GAN paradigm to tabular domains by introducing copula-based transformations that decouple marginal feature modeling from dependency structure~\cite{xu2019modeling}. In particular, features are first mapped into a copula space to facilitate learning of cross-feature correlations, after which a Generator network synthesizes candidate samples and a Discriminator network learns to distinguish real from synthetic instances. Through adversarial optimization, the Generator progressively improves its ability to reproduce both per-feature distributions and multivariate dependencies, yielding higher-fidelity synthetic data for downstream augmentation and evaluation. CopulaGAN was implemented with default parameters using SVD framework. The metadata detection, training, and generation processes mirrored those used for VAE, maintaining consistency in the experimental pipeline. The model was trained with 300 epochs.

\noindent
\textbf{Denoising Autoencoder}. We implemented the denoising autoencoder model using TensorFlow Keras with scikit-learn preprocessing utilities. Numeric features were standardization via StandardScaler, while categorical features were transformed through one-hot encoding. The model compressed the dataset into a lower-dimensional latent representation for subsequent reconstruction~\cite{goodfellow2016deep}, with synthetic data generation achieved through controlled perturbations of latent vectors during decoding.

The architecture comprised a symmetric seven-layer structure with a 32-dimensional bottleneck. The encoder consisted of an input layer matching preprocessed feature dimensionality, a dense layer (64 units, ReLU activation), dropout regularization (rate=0.2), and a bottleneck layer (32 units, ReLU). The decoder mirrored this configuration, terminating in an output layer with sigmoid activation matching the original feature count. This design balanced representational capacity with regularization to prevent overfitting. Training utilized the Adam optimizer with mean squared error loss over 100 epochs (batch size=32). Synthetic sample generation followed a three-stage process: (i) encoding original data into latent representations, (ii) injecting Gaussian noise ($\mu$=0, $\sigma$=0.1) into latent vectors, and (iii) decoding perturbed representations through the reconstruction pathway. This controlled noise injection preserved underlying distributional characteristics while introducing sufficient variation to generate diverse synthetic samples, balancing fidelity with novelty in the generated data. 

\subsection{Evaluation Metrics}
The evaluation metrics that we used to evaluate our models performance are described in this section. 

\noindent
\textbf{Kolmogorov-Smirnov (KS) Test}. Kolmogorov-Smirnov (KS) Test \cite{granville_statistical_2024} measures the maximum distance between the cumulative distribution functions (CDFs) of the real and synthetic data for each feature. A lower KS statistic indicates better similarity between the distributions. If a p-value is included and is less than alpha (generally $0.05$), then we reject the null hypothesis and conclude that the distributions are different. If we do not reject the null, then the distributions are similar. High p-values suggest that the synthetic data closely match the real data for these features. 

The Kolmogorov-Smirnov (KS) test equation is shown in the following:
\begin{equation}
D_{n,m} = \sup_x \left| F_n(x) - G_m(x) \right|
\end{equation}
where $F_n(x)$ and $G_m(x)$ are the empirical CDFs of the real data (of size $n$) and synthetic data (of size $m$), respectively. The test evaluates the null hypothesis $H_0\!: F(x) = G(x)$, which posits that both samples are drawn from the same distribution. A lower KS statistic indicates greater distributional similarity. The null hypothesis is rejected at a significance level $\alpha = 0.05$ if the p-value is less than $0.05$, indicating significant distributional differences between real and synthetic data.

\noindent
\textbf{Jensen-Shannon Divergence}.
The Jensen-Shannon Divergence (JSD) is a symmetric and smoothed version of the Kullback-Leibler (KL) divergence that measures the similarity between two probability distributions. Given two discrete distributions $P$ (distribution of original data) and $Q$ (distribution of synthetic data), the JSD is defined as:

\begin{equation}
\mathrm{JSD}(P allel Q) = \frac{1}{2} \mathrm{KL}(P allel M) + \frac{1}{2} \mathrm{KL}(Q allel M)
\end{equation}

\[
\text{where} \quad M = \frac{1}{2}(P + Q)
\]

Here, $\mathrm{KL}(P allel M)$ denotes the KL divergence between $P$ and the average distribution $M$. The JSD value lies between $0$ (identical distributions) and $1$ (maximum divergence for distributions over two outcomes). A lower JSD indicates greater similarity between the real and synthetic distributions.

\noindent
\textbf{Wasserstein Distance}.

Wasserstein Distance, also known as the Earth Mover\textquotesingle s Distance (EMD), measures the minimum ``cost'' required to transform one probability distribution into another is defined as:

\begin{equation}
  W_p(\mu, \nu) = \left( \inf_{\gamma \in \Gamma(\mu, \nu)} \int_{M \times M} d(x, y)^p \, d\gamma(x, y) \right)^{1/p}  
\end{equation}

\noindent
\begin{itemize}
    \item $p \geq 1$ is the order of the Wasserstein distance.
    \item $d(x, y)$ is the distance between points $x$ and $y$ in the metric space $(M, d)$.
    \item $\Gamma(\mu, \nu)$ denotes the set of all couplings (i.e., joint distributions) $\gamma$ on $M \times M$ with marginals $\mu$ and $\nu$.
\end{itemize}

\noindent
\textbf{ML Model Utility Mean Squared Error and R Squared}.

The mean square error (MSE) is a measure of the average square difference between the actual observed results and the predicted results by the model. It gives an idea of how close the predicted values are to the actual values. Lower MSE values indicate better model performance.

\begin{equation}
\text{MSE} = \frac{1}{n} \sum_{i=1}^{n} (y_i - \hat{y}_i)^2
\end{equation}

Here, $n$ is the number of observations, $y_i$ is the actual value, and $\hat{y}_i$ is the predicted value.

R-Squared (also known as the coefficient of determination) is a statistical measure that represents the proportion of the variance for a dependent variable that\textquotesingle s explained by an independent variable or variables in a regression model. An R-Squared value closer to 1 indicates that the model explains a large portion of the variance in the outcome variable.

\begin{equation}
R^2 = 1 - \frac{\sum_{i=1}^{n}(y_i - \hat{y}_i)^2}{\sum_{i=1}^{n}(y_i - \bar{y})^2}
\end{equation}

In this formula, $\bar{y}$ is the mean of the actual values. The numerator represents the residual sum of squares (RSS), and the denominator represents the total sum of squares (TSS). A higher $R^2$ value indicates a better fit of the model to the data.

\noindent
\textbf{Train-on-Synthetic-Test-on-Real (TSTR)}.
The Train on Synthetic, Test on Real (TSTR) score quantifies the utility of synthetic data for predictive modeling by assessing model generalization from synthetic training data to real-world test data. This metric operates on the principle that high-quality synthetic data should enable models to learn transferable patterns that generalize effectively to authentic data distributions. We evaluated TSTR performance across multiple dimensions: (a) Classification TSTR employed a Random Forest classifier with 200 estimators, measuring accuracy on a 30\% held-out real data test set after training exclusively on synthetic samples; (b) Regression TSTR utilized a Random Forest regressor with 300 estimators, evaluating both R-squared and Mean Absolute Error metrics following the same train-synthetic/test-real protocol; and (c) ML Utility Aggregate score computed a composite metric combining classification accuracy and regression R-squared values to provide a holistic assessment of synthetic data quality across both predictive tasks.

\begin{table*}[t]
\centering
\footnotesize
\caption{Evaluation metrics for synthetic data generation methods. Lower values (↓) indicate better distributional similarity; higher values (↑) indicate superior performance. 
}
\label{tab:synth_eval_metrics}
\resizebox{\textwidth}{!}{
\begin{tabular}{@{}lccccccc@{}}
\textbf{Method} & 
\textbf{KS} & 
\textbf{Wasserstein} & 
\textbf{JS} & 
\textbf{Train-on-} & 
\textbf{Cat.} & 
\textbf{Distance} & 
\textbf{ML } \\
& \textbf{Distance ↓} & \textbf{Distance ↓} & \textbf{Divergence ↓} & \textbf{Synth ↑} & \textbf{Fidelity ↑} & \textbf{to Closest ↑} & \textbf{Utility ↑}\\
\hline
SMOTE        & 0.10 & 3.57  & 0.04 & 0.997 & 0.998 & 0.02 & 0.92 \\ 
Bootstrap    & 0.11 & 4.07  & 0.04 & 0.997 & 0.993 & 0.00 & 0.92 \\ 
Oversampling & 0.10 & 3.68  & 0.04 & 0.997 & 0.993 & 0.00 & 0.92 \\ 
Autoencoder  & 0.07 & 2.51  & 0.08 & 0.533 & 0.982 & 0.99 & 0.67 \\ 
VAE          & 0.13 & 5.46  & 0.06 & 0.833 & 0.923 & 1.00 & 0.58 \\ 
CopulaGAN    & 0.34 & 15.82 & 0.18 & 0.700 & 0.992 & 1.00 & 0.15 \\ 
\hline
\end{tabular}
}
\end{table*}

\noindent
\textbf{Categorical Fidelity}.
This measures how well the synthetic data preserves the distribution of categorical variables from the original dataset. It ensures that the frequency and diversity of categories (e.g., gender, region, product type) are realistically replicated. Example: If the original data has 60 percent ``Yes" and 40 percent ``No" in a column, high categorical fidelity means the synthetic version closely mirrors that ratio.

\noindent
\textbf{Distance to Closest Record}.
Distance to Closest Record (DCR) is a key privacy metric. For each record in a synthetic dataset, this metric finds the single closest matching record in the real dataset. ``Mean DCR" is simply the average of all these distances.
A higher mean DCR is better. It suggests that the synthetic records are not just simple copies of the real data, which is good for privacy. A very low DCR would mean that synthetic data is too similar to the original, creating a privacy risk.

\section{Findings}
\begin{figure}[!t]
    \centering
    \begin{minipage}[b]{0.32\textwidth}
        \centering
        \includegraphics[width=\linewidth,height=0.14\textheight]{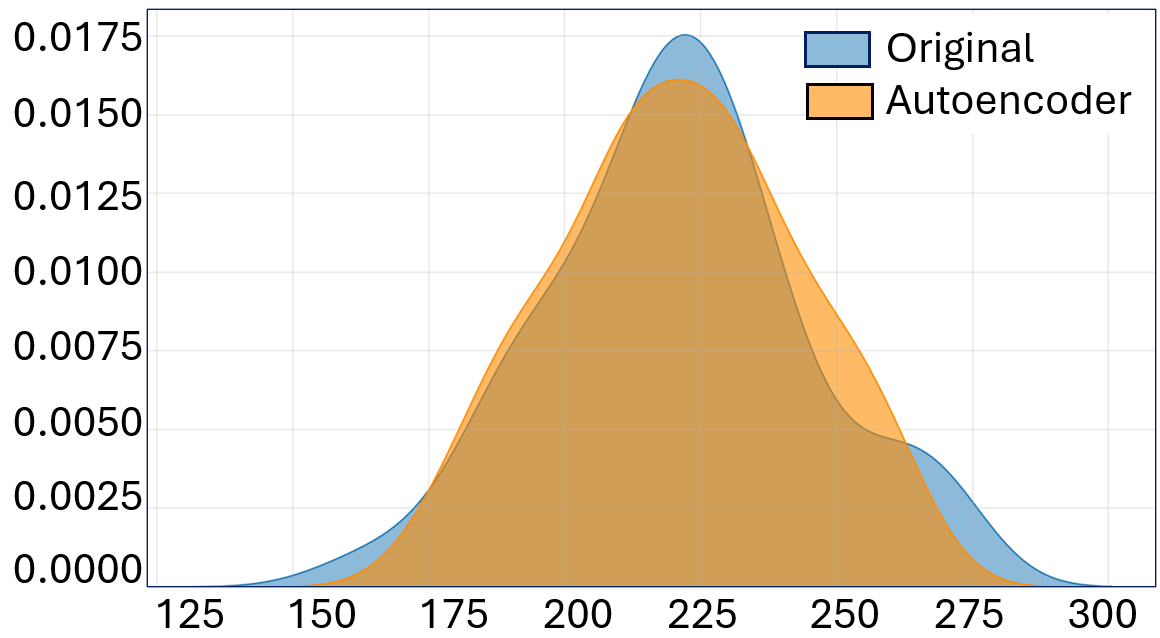}
        \textbf{Autoencoder}
    \end{minipage}
    \hfill
    \begin{minipage}[b]{0.32\textwidth}
        \centering
        \includegraphics[width=\linewidth,height=0.14\textheight]{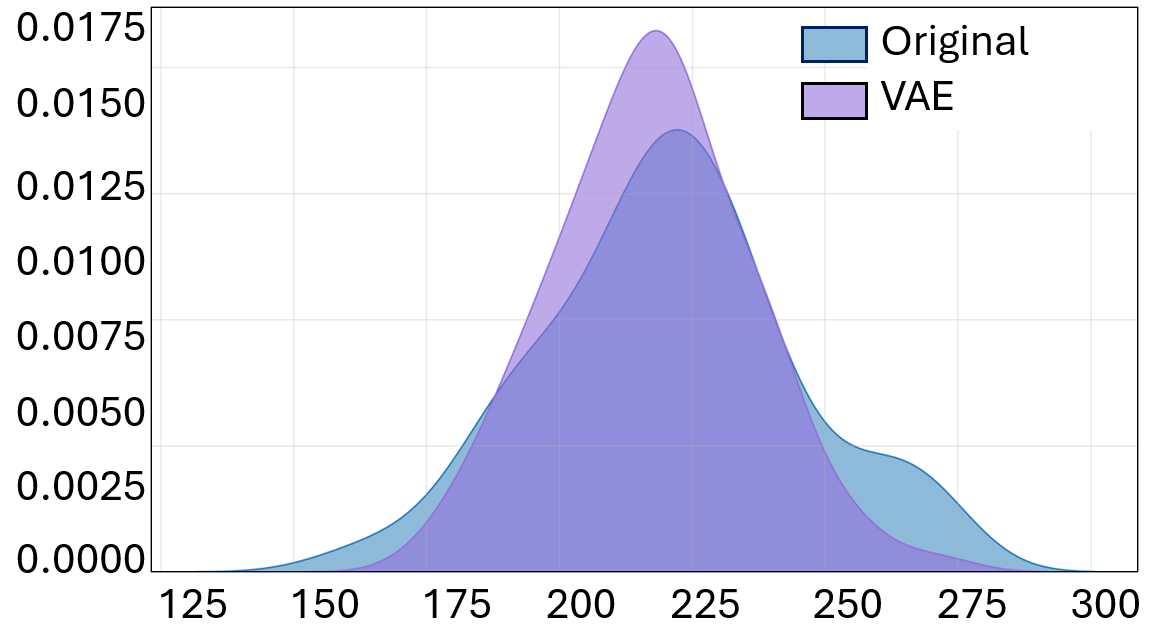}
        \textbf{VAE}
    \end{minipage}
    \hfill
    \begin{minipage}[b]{0.32\textwidth}
        \centering
        \includegraphics[width=\linewidth,height=0.14\textheight]{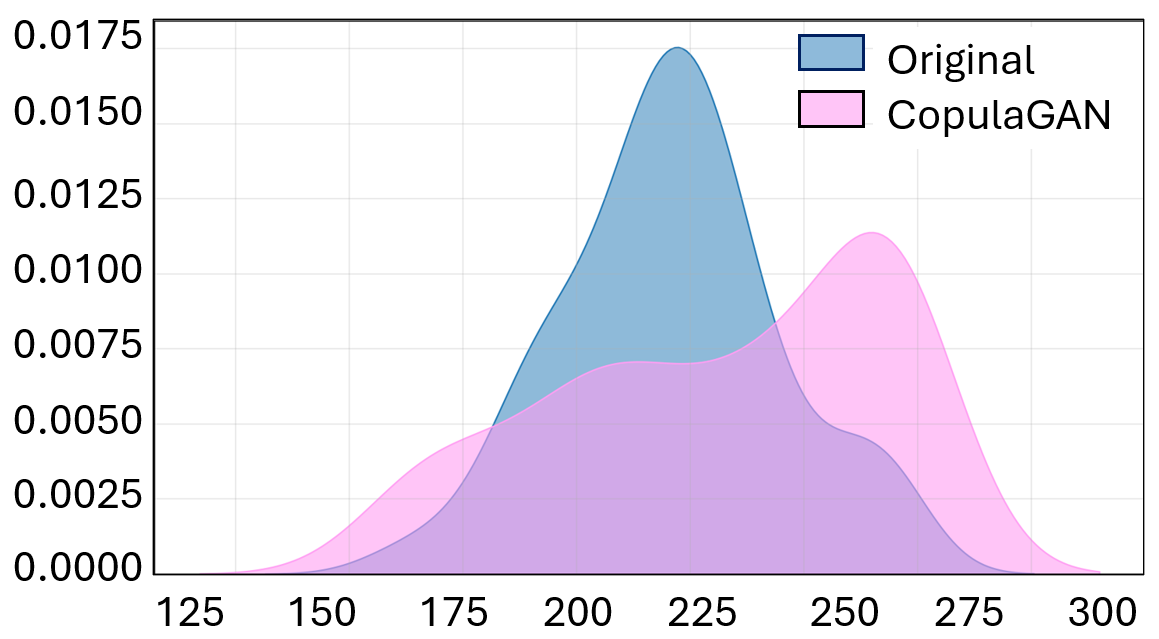}
        \textbf{CopulaGAN}
    \end{minipage}
    \caption{KDE plots comparing total score distributions between original data and synthetic datasets generated by three deep-learning methods. The x-axis indicates the total test scores and y-axis is the density. }
    \label{fig:kde_comparison}
\end{figure}

Table~\ref{tab:synth_eval_metrics} summarizes the quantitative evaluation, while Figure~\ref{fig:kde_comparison} provides a qualitative KDE-based comparison across three synthetic data generation methods, each producing 10,000 synthetic records. Together, these results highlight a fundamental utility–privacy tension: approaches that more closely match the original distribution tend to yield higher downstream utility, but may also retain more identifiable structure, whereas methods that introduce stronger perturbations can improve privacy at the cost of fidelity. The evaluated techniques naturally partition into two families with distinct operational logics traditional resampling methods, which rebalance the dataset through duplication or local interpolation of real instances, and deep generative models, which learn an explicit approximation to the joint data distribution and sample novel records from the learned manifold.

\subsection{Traditional Resampling Methods}

SMOTE, Bootstrap, and other classical oversampling techniques exhibit exceptionally strong utility but fail severely with respect to privacy. These methods achieve near-perfect distributional fidelity (KS: 0.10–0.11; JS: 0.04) and train-on-synthetic–test-on-real (TSTR) performance of 0.997, indicating that models trained solely on synthetic data recover 99.7\% of the predictive performance attainable with real data. Their overall ML utility score of 0.92 further confirms their high practical value for downstream modeling tasks.

Despite these advantages, their privacy performance is fundamentally deficient. A distance-to-closest-record score of almost zero indicates complete memorization, meaning synthetic samples directly replicate instances in the training set and provide no meaningful privacy protection. As a result, these methods are appropriate only for internal machine learning workflows in secure environments such as class-imbalance correction, data augmentation, or rapid prototyping, and are entirely unsuitable for external data release or any context requiring privacy preservation.

\subsection{Generative Models}

Deep learning approaches offer privacy protection (distance-to-closest: 1.00) but at significant utility cost.

\noindent
\textbf{Autoencoder}.
Despite excellent distributional distances (KS: 0.07, Wasserstein: 2.51), autoencoders suffer severe ML utility degradation (train-on-synthetic: 0.533). This paradox indicates over-smoothing: preserving approximate distributions while losing critical high-frequency information. With nearly perfect privacy protection (DCR .99) but limited utility (0.67), autoencoders suit maximum-privacy scenarios where utility degradation is acceptable.

\noindent
\textbf{Variational Autoencoder}.
VAEs achieve optimal privacy-utility balance among generative methods: perfect privacy (1.00) with train-on-synthetic performance of 0.833, a 30-percentage-point improvement over standard autoencoders. The probabilistic framework with KL divergence regularization prevents memorization while maintaining 83.3\% of real-data ML performance. VAEs represent the preferred choice for privacy-sensitive external sharing in healthcare, finance, and regulatory compliance scenarios.

\noindent
\textbf{CopulaGAN}.
CopulaGAN demonstrates catastrophic failure for tabular data (ML utility: 0.15, Wasserstein: 15.82) despite perfect privacy. The adversarial framework fails with heterogeneous tabular features due to discriminator inadequacy with discrete features, training instability, and mode collapse \cite{choi2017generating, xu2019modeling}. The paradoxical high categorical fidelity (0.992) alongside poor continuous feature modeling indicates fundamental architectural mismatch. 

\section{Conclusion and Future Work}

This study provides the first comprehensive benchmark comparing traditional resampling techniques against deep learning approaches for synthetic educational data generation, revealing fundamental trade-offs with critical practical implications. Our empirical evaluation demonstrates that no single method dominates across all metrics. Traditional resampling techniques (SMOTE, Bootstrap, Random Oversampling) achieve 99.7\% of real-data predictive performance but completely fail privacy protection (DCR $\approx$ 0.00), essentially producing copies of training data. Conversely, deep learning models guarantee privacy (DCR $\approx$ 1.00) with varying utility costs: autoencoders retain 53.3\% performance, VAEs maintain 83.3\%, while Copula-GANs fail for heterogeneous tabular data (15\% performance). These findings show clear practical recommendations: organizations should employ traditional resampling for internal development where maximum utility is essential, but use Variational Autoencoders for external data sharing or privacy-sensitive applications. This strategic separation enables high-performance internal models while ensuring regulatory compliance.

In future we will investigate: (1) adaptive frameworks that dynamically adjust generation methods based on privacy-utility requirements, (2) domain-specific architectures optimized for educational data patterns, (3) formal privacy guarantees beyond empirical distance metrics, and (4) hybrid approaches combining resampling with differential privacy. Validation across diverse educational datasets and extension to sequential learning traces remain important open challenges. As educational institutions increasingly balance data-driven insights with privacy regulations, this benchmark provides essential guidance for responsible synthetic data deployment.

\section*{Acknowledgements} \label{Acknowledgements}
The authors acknowledge the use of generative AI tools to assist in improving the fluency and grammar of the manuscript. No content was generated beyond these editorial refinements.

\bibliographystyle{splncs04}
\bibliography{references}
\end{document}